\documentclass[preprint,12pt]{elsarticle}

\usepackage{graphicx}
\usepackage{amsmath}
\usepackage{amssymb}
\usepackage{booktabs}
\usepackage{colortbl}

\usepackage[margin=1in]{geometry}
\usepackage[table]{xcolor}
\usepackage{tikz}
\usetikzlibrary{shapes,arrows}
\usetikzlibrary{positioning,decorations.pathreplacing}
\usepackage{amsmath,bm}
\usepackage{amssymb}
\usetikzlibrary{positioning}
\usepackage{subcaption}
\usepackage{algorithm,algpseudocode}
\usepackage{mathtools}
\usepackage{rotating,caption,booktabs}
\usepackage{float}
\usepackage{hvfloat}
\usepackage{booktabs}
\usepackage{adjustbox}
\usepackage[hidelinks]{hyperref}
\usepackage{natbib}
\setcitestyle{authoryear,open={(},close={)}}

\usepackage[normalem]{ulem}
\usepackage{makecell, multirow}
\usepackage{scrextend}
\usepackage{footnote}
\makesavenoteenv{tabular}
\makesavenoteenv{table}
\usepackage{bbold}
\usepackage{lscape}
\usepackage{lineno}

\usepackage{titlesec}
\usepackage{lmodern}
\titleformat*{\subsection}{\fontsize{13}{18}\itshape}

\makeatletter
\def\ps@pprintTitle{%
 \let\@oddhead\@empty
 \let\@evenhead\@empty
 \def\@oddfoot{\centerline{\thepage}}%
 \let\@evenfoot\@oddfoot}
\makeatother

\usepackage[capitalize]{cleveref}
\crefname{section}{Sec.}{Secs.}
\Crefname{section}{Section}{Sections}
\Crefname{table}{Table}{Tables}
\crefname{table}{Tab.}{Tabs.}

\begin{document}

\begin{frontmatter}

\title{Images in Discrete Choice Modeling:\\ Addressing Data Isomorphism in Multi-Modality Inputs}

\author[epfl]{Brian Sifringer}
\author[epfl]{Alexandre Alahi}
\address[epfl]{Visual Intelligence for Transportation Laboratory (VITA), \'Ecole Polytechnique F\'ed\'erale de Lausanne (EPFL), Switzerland.}

\begin{abstract}
 
This paper explores the intersection of Discrete Choice Modeling (DCM) and machine learning, focusing on the integration of image data into DCM's utility functions and its impact on model interpretability. We investigate the consequences of embedding high-dimensional image data that shares isomorphic information with traditional tabular inputs within a DCM framework. Our study reveals that neural network (NN) components learn and replicate tabular variable representations from images when co-occurrences exist, thereby compromising the interpretability of DCM parameters. We propose and benchmark two methodologies to address this challenge: architectural design adjustments to segregate redundant information, and isomorphic information mitigation through source information masking and inpainting. Our experiments, conducted on a semi-synthetic dataset, demonstrate that while architectural modifications prove inconclusive, direct mitigation at the data source shows to be a more effective strategy in maintaining the integrity of DCM's interpretable parameters. The paper concludes with insights into the applicability of our findings in real-world settings and discusses the implications for future research in hybrid modeling that combines complex data modalities. Full control of tabular and image data congruence is attained by using the MIT moral machine dataset, and both inputs are merged into a choice model by deploying the Learning Multinomial Logit (L-MNL) framework.

\end{abstract}


\end{frontmatter}

\clearpage

\section{Introduction}
\label{sec:intro}

In recent years, the proliferation of Deep Learning (DL) and Neural Networks (NN) has been remarkable, driven by the increasing availability of big data and advancements in computing power. This surge has seen DL and NNs enter in practically all scientific fields, including Discrete Choice Modeling (DCM) \citep{hillel2021systematic,van2022choice}. A key strength of NNs lies in their versatility to process and learn from diverse data types, ranging from structured tabular or sequential data to high-dimensional formats like images, video, or natural language processing. As these varied data modalities often hold valuable insights for decision-making processes, their integration into DCM is not just imminent but also essential. Indeed, using visual data in transportation, for example, is already a growing field \citep{ramirez2021measuring,basu2022street,ito2021assessing,juarez2023cyclists}.

A significant challenge in creating hybrid NN-DCM models, however, is maintaining the interpretability that is central to the appeal of DCM. Recent studies have tackled this challenge using tabular data \citep{sifringer_enhancing_2020,han2022neural,lahoz2023attitudes}, but the integration of high-dimensional data into DCM is a relatively new endeavor \citep{van2023computer,szep2023moral}. We believe the potential for high correlation or redundancy between different data modalities, particularly in the context of DCM, poses a set of complex questions and necessitates thorough investigation to ensure the integrity of interpretable hybrid models. In our study we specifically focus on the integration of visual data into DCM models, examining the implications of this convergence.
Where initial studies employing neural networks (NN) to embed image data into DCM utility functions have seen successful results \citep{van2020blending,van2023computer}, the implications of merging visual information with standard tabular inputs—notably the potential redundancy between these modalities—remain an open question.

\begin{figure}[H]
    \centering
    \includegraphics[width=\textwidth, trim={0 8.5cm 0 0},clip]{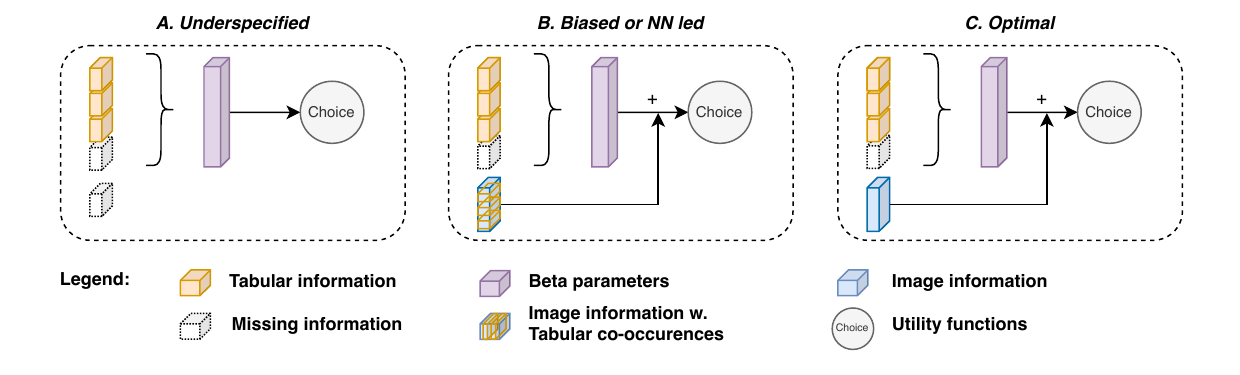}
    \caption{Using only tabular data when multiple modalities are available could lead to a sub-optimal model (A). When data isomorphism between two modalities exists, co-occurring information may lead to biased parameters (B). An optimal model requires both modalities to contain solely independent information (C).}
    \label{fig:pull_figure}
\end{figure}

Our study addresses this gap by examining the impact of overlapping information between image and tabular data on the interpretability of DCM parameters. We present two key contributions: firstly, we demonstrate that an NN's ability to extract latent representations of tabular variables from images can compromise the interpretability of DCM. Secondly, we investigate two ways to handle this challenge and bring to light the best approach. Initially, we explore architectural designs and modeling techniques to try to segregate redundant information within the network in order to preserve the model's statistical interpretability. We then present our solution of isomorphic information mitigation at the source. This consists of first detecting and then removing redundant data directly from the inputs, which allows the model to learn and incorporate only the complementary information into the DCM. A simple description motivating the challenge posed by data information co-occurrences for a hybrid DCM model can be seen in Figure (\ref{fig:pull_figure}).

The paper is organized as follows: section (\ref{sec:literature}) will go through relevant literature, section (\ref{sec:methodology}) will shortly describe methodology used to mix image and tabular datasets in DCM as well as our approach to mitigate redundant information at source, section (\ref{sec:experiments}) will contain all relevant experiments conducted and finally section (\ref{sec:conclusions}) will summarize how to approach combined image and tabular data for DCM and give real-world examples applications to our study. All our code is openly available to promote open science.\footnote{Code available at: \url{https://github.com/BSifringer/ImageDCM}} \\ 

\section{Literature}\label{sec:literature}

Discrete Choice Modeling (DCM) is highly prevalent in the fields of Transportation, Economy, and Health \citep{bierlaire1998discrete, haghani2021landscape,soekhai2019discrete}. The appeal of DCM models lies not only in their interpretability and mathematical foundations but also in their use to forecast decision changes based on varying features. This allows for instance to understand behavioral shifts, such as preferences in travel modes or consumer products. In recent years, acute interest in applying more complex machine learning methods to these fields has arisen as observed in the following reviews for transportation \citep{hillel2021systematic,van2022choice}.\\

While advancements in machine learning have shown high predictive accuracy in decision-modeling tasks, the preservation of interpretability, which is invaluable in decision-making contexts, is often overlooked. There are quite a few recent works that have studied this direction. Some works show the possibility of recovering interpretability by analyzing the distributions from neural network (NN) components \citep{wang2020deepa, wong_bi-partite_2020}, by preserving the model's original structure \citep{sifringer_enhancing_2020, arkoudi2023combining, han2022neural, phan2022attentionchoice, van2023computer, lahoz2023attitudes}, or by employing a combination of these strategies \citep{wang2020deepb, wang2021theory, szep2023moral}. Our research aims to retain the core structure of DCM while augmenting it with image-derived information, aligning our approach with the second strategy. Among these works, in \cite{han2022neural}, they learn systematic taste heterogeneity using a NN while keeping interpretability on other regular DCM parameters. This is to take into account that not all individuals apply the same value of importance for different attributes for a given choice. An improved model following their framework is proposed by \cite{phan2022attentionchoice}, using an attention mechanism for the NN component. In \cite{arkoudi2023combining}, they leverage the contextual embedding properties of NN for both: an improved interpretability in the neural network component as well as added regularization when using sparse variable occurrences. \cite{lahoz2023attitudes} successfully extend the use of NN for hybrid interpretable DCM in attitudes and Latent Class choice models. Finally, our work resembles most \cite{sifringer_enhancing_2020} and \cite{van2023computer} presented here and in the next literature topic.
In \cite{sifringer_enhancing_2020} they have a NN infer a data-driven representation term for the utility function, by taking data that is not used by the original DCM model. This adds information missed during the modeling phase, improving predictability and potentially fixing for utility misspecification. Given we wish to add information to the utility with an image and not use it to estimate parameter heterogeneity, we base our work on that of \cite{sifringer_enhancing_2020}.\\

Adding complex data into a utility function such as an image has only been done by a few papers as of our knowledge.
Preliminary works in this field by \cite{van2020blending} conducted a synthetic experiment demonstrating how a NN can rank images by aesthetics for decision-making purposes. Their most recent work, as cited above \citep{van2023computer}, is co-occurring and complementary to ours. They define a new model, computer vision-enriched discrete choice models (CV-DCM), in which they show the benefits of adding image information directly to the utility function using both image and tabular data. The paper is the first to our knowledge to show these positive results. Our work complements this approach as we investigate when these models could instead bias the interpretable parameters. Moreover, both our findings synergize well and are cross-compatible, as their CV-DCM model is a particular case within the L-MNL framework. Indeed, the neural network terms arise from representations of a complementary input set and are mixed into the utility function by straightforward addition. While we also propose an image-based neural network component for the residual component in L-MNL, we do not find a particular relevance to give a new model name within this framework.

Another study worth noting has shown the feasibility of learning decision-based predictions using satellite images \citep{wang2023deep}. Their model is trained with the task of generating images while predicting their associated socioeconomic information. While the approach presents a novel way to use both modalities, tabular and image, in tasks of choice modeling, the use of simple utility functions and having straightforward interpretable parameters is left aside in favor of a generative model for recreating insightful data distribution. \\

Finally, in our research, we introduce an image-derived embedded term into a DCM framework. The DCM field encompasses a variety of theoretical models, with the Multinomial Logit (MNL) \citep{hausman1984specification} being the most prevalent, based on the theory of Random Utility Maximization \citep{manski1977structure}. Although there are numerous other models tailored to specific tasks, datasets, and theoretical assumptions \citep{williams1977formation,mcfadden1978modeling,vij2013,shen2009}, our focus on MNL allows for clear observation of our findings, with implications that can be extrapolated to other models. As such, we do not go in further details for this review.

\section{Methodology}\label{sec:methodology}
\subsection{Models}

In this paper, we utilize only two core models: the Multinomial Logit (MNL) \citep{hausman1984specification} and the Learning Multinomial Logit (L-MNL) \citep{sifringer_enhancing_2020}. The former is adopted as the representative of the most common and straightforward statistical model in DCM, making it well-suited for our experimental benchmarks. The L-MNL, on the other hand, is the only model that allows a direct comparison between the MNL weights and its own interpretable parameters while incorporating an image into its input. It's worth noting that we will investigate a variety of baselines, all of which will stem from these two models, differing only in terms of a loss function or regularizer.

Without going too deep into DCM theory, we adopt the standard MNL assumptions. For a choice set $\mathcal{C}$ and inputs $\mathcal{X}_n$ of individual $n$, the utility function can be expressed as:
\begin{equation}\label{eq:utility}
     \mathbf{U}_n = \mathbf{f}(\mathcal{X}_n,\bm{\beta}) + \mathbf{\varepsilon}_n ,
\end{equation}
where $\bm{\beta}$ represents the interpretable weights, $\varepsilon_{in} \overset{\text{i.i.d.}}{\sim} EV(0, 1)$ $\forall i \in \mathcal{C}\), and $f(\cdot)$ is chosen to be linear in parameters.

Similarly, the utility of L-MNL can be expressed as:
\begin{equation} \label{eq:ourutility}
  \mathbf{U}_{n} =  \mathbf{f}(\mathcal{X}_n;\bm{\beta}) + \mathbf{r}(\mathcal{I}_n;\mathbf{w}) + \mathbf{\varepsilon}_n ,
\end{equation}
where $\mathcal{I}_n$ is the image associated with individual $n$'s choice and $\mathbf{w}$ represents the weights in the deployed neural network. It's important to note that we can decompose $r(\cdot)$ into two steps. First, an image model $B(\cdot)$ is used as a backbone to create a latent space $\mathbf{l}_n$, and then a head $H(\cdot)$ is utilized to condense the information of $\mathbf{l}_n$ into a single term per utility $r_{in}$ $\forall i \in \mathcal{C}\). This can be expressed as:
\begin{equation}
 \mathbf{r}(\mathcal{I}_n; \mathbf{w}) = {H}(\mathbf{l}_n, \mathbf{w}_h) = {H}({B}(\mathcal{I}_n,\mathbf{w}_b), \mathbf{w}_h),
\end{equation}
where $\mathbf{w}_h$ and $\mathbf{w}_b$ are the weights in components $H(\cdot)$ and $B(\cdot)$ respectively. In our study, $B(\cdot)$ was chosen to be the ResNet18 \citep{he2016deep}, and $H(\cdot)$ is a simple linear layer by default. Note that the last layer of ResNet18 was removed in order to use it as a backbone and access the latent space of $512$ features.\\

Finally, the probability of choosing utility $i$ for individual $n$ can be written as:
\begin{equation} \label{eq:softmax}
P(U_{in}|\mathcal{C}) =  \frac{e^{V_{in}}}{\sum_{j\in \mathcal{C}} e^{V_{jn}}} =  (\bm{\sigma}(\textbf{V}_n))_i  ,
\end{equation}
where $V_{in}$ is the deterministic part of $U_{in}$ and $\sigma(\cdot)$ is the softmax function.\\

\begin{figure}
    \centering
      \includegraphics[width=0.8\textwidth, trim={0 2cm 2cm 0},clip]{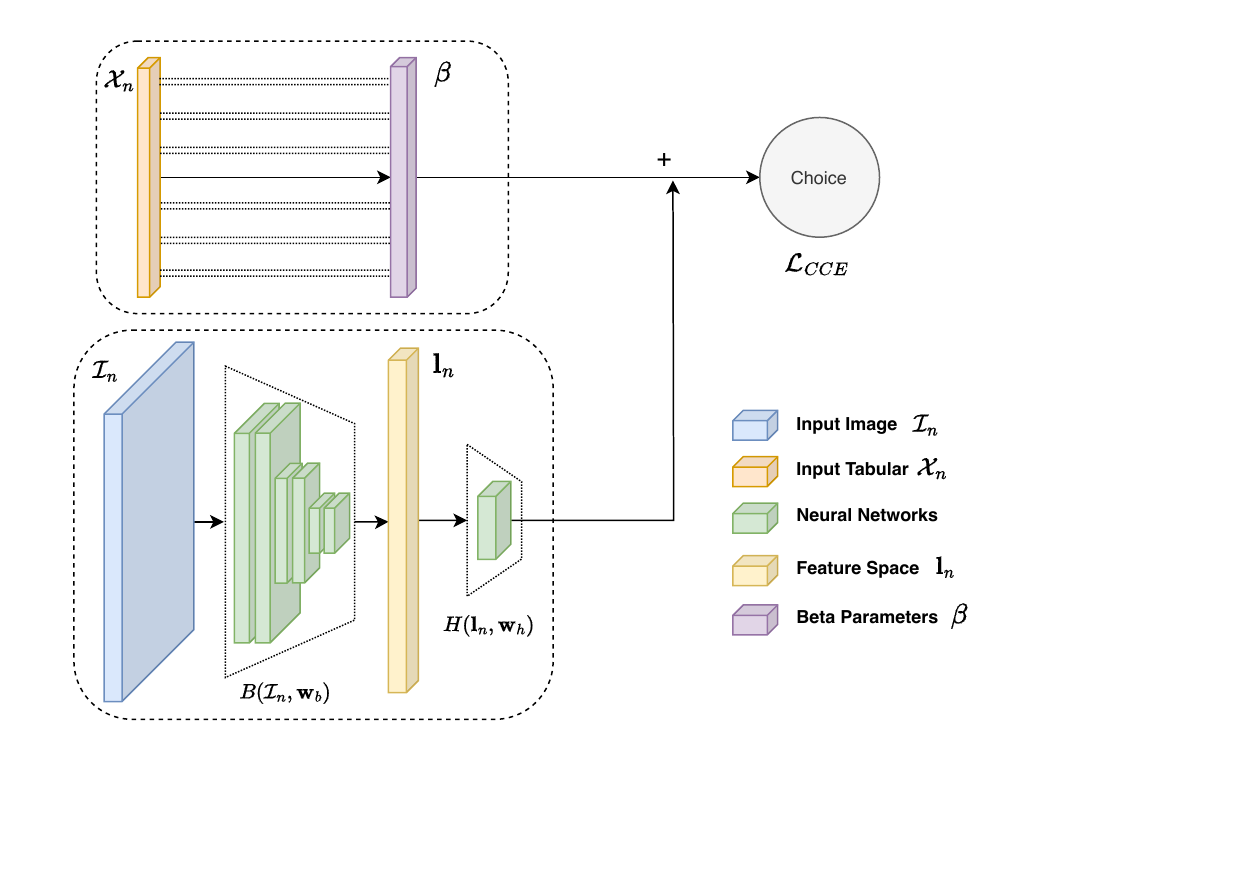}
    \caption{L-MNL model with MNL component and Image neural network concatenated with a sum.}
    \label{fig:lmnl}
\end{figure}

The default loss used to train the models for choice prediction is the average negative log-likelihood, or categorical cross-entropy loss, written as:
\begin{equation}
    \mathcal{L}_{CCE} = -\frac{1}{N}\sum_{n=1}^N\sum_{i \in \mathcal{C}} y_i \log(P(U_{in}|\mathcal{C})) ,
\end{equation}
which is then summed for all individuals n. An illustration of the model may be observed in Figure (\ref{fig:lmnl}).

\subsection{Isomorphic Information Mitigation}\label{sec:methodology_mitigate}

In this section, we introduce our proposed methodology to ensure the preservation of interpretable parameters in our statistical model when isomorphic information exists between tabular and image data. As we will demonstrate in subsequent experiments, the presence of image information can adversely influence the statistical model (section \ref{sec:initial}). Attempts to architecturally control the information flow will be proven inadequate, with the Neural Network often overshadowing the statistical model (section \ref{sec:baselines}). Our proposed solution is to disrupt the source information in the Neural Network, preventing it from accessing isomorphic information present in the tabular data.

To achieve this, we employ a detection model, denoted as $M$, leveraging the saliency capabilities inherent in Neural Networks. The objective is to identify regions in the original image, $\mathcal{I}_n$, that contain redundant information. This is accomplished by feeding the image into the model and training it to reconstruct the corresponding tabular data, $\mathcal{X}_n$. Depending on the specific data, model, and experimental context, an appropriate reconstruction loss, $\mathcal{L}_R$, can be selected. This relationship is expressed as:
\begin{equation} \label{eq:detect_tab}
\mathcal{L}_R := \mathcal{L}_R(M(\mathcal{I}_n), \mathcal{X}_n) .
\end{equation}
By employing gradient visualization techniques, we generate a heatmap for each image, highlighting regions that correspond to elements in the tabular input. Using this heatmap and a threshold function, we can create a mask $\mathcal{M}_n$ over the image. The identified regions can either be erased (set to zero) or inpainted. We refer to the former method as ``black masking", transforming $\mathcal{I}_n$ to $\mathcal{I}_{nb}$, and the latter as ``background inpainting," converting $\mathcal{I}_n$ to $\mathcal{I}_{np}$. The black masking operation may be written as: 
\begin{equation}
    \mathcal{I}_{nb} = \mathcal{I}_n \odot (\mathbb{1}-\mathcal{M}_n) ,
\end{equation}
for a mask $\mathcal{M}_n$ with values either 0 or 1. The inpainting can be done with an appropriate model we denote $P(\cdot)$ such that:
\begin{equation}
    \mathcal{I}_{np} = P(\mathcal{I}_{nb}) .
\end{equation}

Numerous models in the literature address image classification \citep{rawat2017deep, chen2021review}, object detection \citep{liu2020deep, sharma2020comprehensive}, gradient visualization methods \citep{zhang2018visual, burkart2021survey}, and inpainting \citep{elharrouss2020image, jam2021comprehensive}. The choice for each component of this multi-step system depends on the data format, choice sets, and tasks at hand. In our experiments, we easily achieve 100$\%$ detection accuracy using the aforementioned ResNet18 backbone, denoted as $B(\cdot)$, combined with a straightforward linear detection head, $D(\cdot)$, given the clear synthetic setting we operate within. As elaborated further in experiment section \ref{sec:exp_masking}, both masking and inpainting are executed algorithmically. This approach allows us to comprehensively examine the effects stemming from varying qualities of source image disruption. The rationale behind employing synthetic data for this study, along with its applicability to real-world scenarios, is discussed in section \ref{sec:real_world}.

\section{Experiments}\label{sec:experiments}

\subsection{Data}\label{sec:data}
Our experiments necessitate tabular input paired with image data within the field of visual decision-making. To fulfill this requirement, we utilize the publicly accessible MIT moral machine dataset \citep{awad_moral_2018}. This dataset is derived from crowd-sourced responses to a series of randomized moral dilemma scenarios. In each scenario, participants decide between two potential outcomes: a vehicle with brake failure either continues straight or swerves. The consequences of each choice determine the fate of various individuals or animals, reminiscent of the well-known trolley problem \citep{thomson1984trolley}. Additionally, participants have the option to complete a supplementary survey that gathers personal details, such as age, income, political inclinations and more.

For the purposes of our study, we refine the dataset in several ways. Initially, we only consider participants who have completed both the primary and secondary surveys. We further exclude respondents who have addressed more than 100 questions to mitigate potential bot interference. From a DCM point of view, the MIT moral machine dataset encompasses two distinct scenario types: choices between a vehicle crashing into pedestrians in both alternatives or decisions where the vehicle either hits pedestrians or collides with a barrier, endangering its passengers. To maintain consistency in our choice modeling, we exclusively focus on the former, retaining only scenarios that present pedestrian-versus-pedestrian dilemmas. From this refined dataset, we sample 80,000 paired entries, equating to 40,000 unique scenarios, as each scenario offers two choice alternatives. We then partition the data into training and testing sets in a 75/25 ratio, while ensuring that responses from individual participants are confined to a single set.

\begin{figure}
    \centering
    \includegraphics[width=0.7\textwidth]{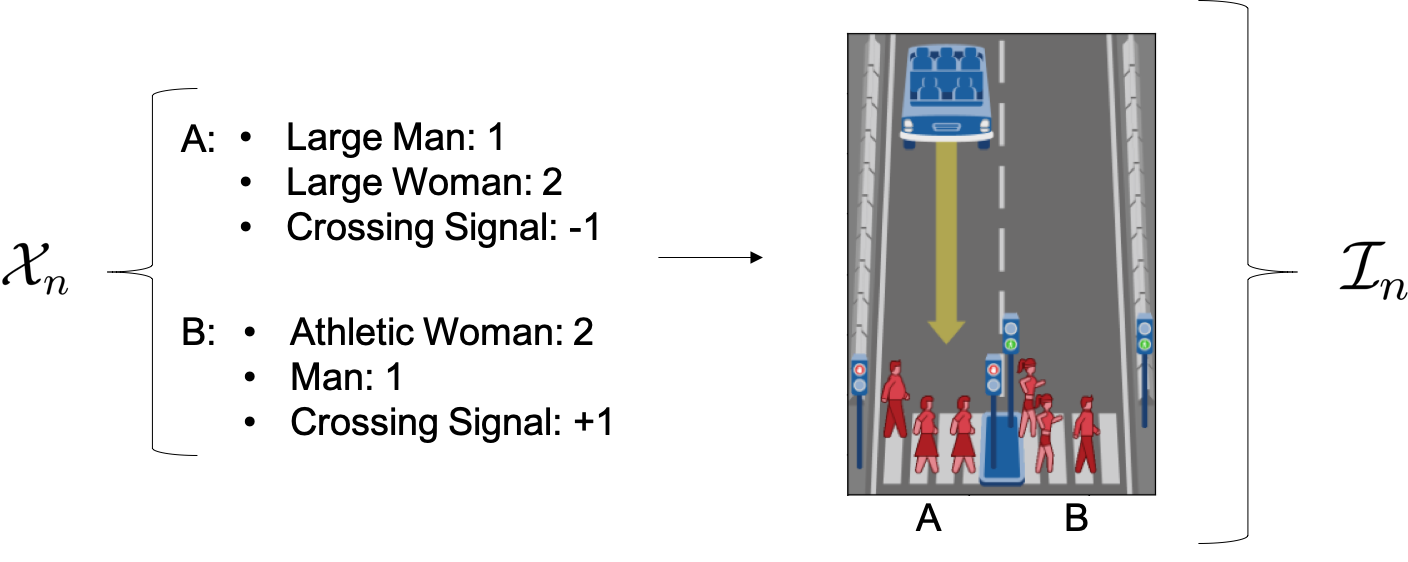}
    \caption{Image reconstruction from Tabular information. The single image contains all necessary information relative to choosing A (‘straight') versus B (‘swerve'). }
    \label{fig:data_convert}
\end{figure}

Although the MIT moral machine dataset originates from image-based scenario selections, it is only stored in tabular format. However, this structure allows us to reconstruct the original images from the tabular data, up to a permutation in character positioning. For each scenario, we produce a single image in which both outcomes may easily be deduced. In this sense, the dataset is real for the tabular information, over which each scene is synthetically reconstructed for the image part. The reconstruction is done thanks to a background image and character tiles we position left or right based on the scenario. An example can be seen in Figure (\ref{fig:data_convert}). 

\subsection{Initial study and setup}\label{sec:initial}

Initially, we conduct a sanity check as detailed in \ref{sec:sanity}, where we compare our predictive performance against established benchmarks from the literature and evaluate the detection accuracy of our selected backbone $B(\cdot)$. Subsequently, we train our core models on our selected data $\mathcal{X}$: MNL(\(\mathcal{X}\)) and L-MNL(\(\mathcal{X},\mathcal{I}\)), aiming to discern shifts in the interpretable parameters $\bm{\beta}$. Note that for all our experiments, we use the Adam optimizer \citep{kingma2014adam} with a learning rate of lr=$10^{-4}$. We also always load the pre-trained weights of the DCM component when training L-MNL, and multiply the learning rate of the beta parameters by 20. This is so that both components reach convergence in a decent amount of time. In our case we train for 15 epochs for all experiments. The outcomes of this training can be observed in Table \ref{tab:1-1}. We may thus observe, due to the direct correlation between the tabular and image inputs, that a conflict arises between the image component and the MNL part during joint loss minimization. Indeed, the neural network component will learn latent representations of the useful variables found on its image, and thus bias the MNL's parameters. We also note that
the ‘intervention' variable represents an essential constant for MNL models, however, its alteration is inevitable since the image model can independently introduce its own constant outcome bias. We therefore systematically remove its comparison for all hybrid models in this paper.

While it is expected that 1-1 isomorphism between both datasets would induce bias in the interpretable parameters for the L-MNL model, we may now examine a more realistic approach with data complementarity.\\

 Throughout the remainder of our study, we modify the tabular input to emulate real-world scenarios where the image conveys more information than its tabular counterpart. To achieve this, we omit a specific component from the tabular data. Specifically, we exclude all instances representing the ‘Man' input, resulting in a new dataset denoted as $\mathcal{X}'$. This omission is designed to simulate utility specifications lacking a crucial variable, leading to potential misspecification and parameter bias. The image data, on the other hand, remains unaltered, representing a comprehensive information set not entirely captured in tabular format. The results for MNL($\mathcal{X}'$) and L-MNL($\mathcal{X}'$, $\mathcal{I}$) are referenced in Table \ref{tab:1-1}. 

A clear observation is that the removal of a significant factor from the tabular data distorts the original $\beta_{MNL}$ parameters, resulting in a highly misspecified variant $\beta_{MNL}'$. This distortion is particularly evident for the ‘OldMan' variable, which approaches a value close to zero. One might anticipate that joint optimization would enable the image model to identify and learn solely the omitted information from $\mathcal{X}'$, thereby restoring the utility to its optimal parameters $\beta_{MNL}$. However, while many parameters in $\beta_{LMNL}'$ are rectified, they continue to exhibit the correlation effects observed in our initial experiment. It is therefore clear that a naive joint optimization will not suffice in avoiding parameter bias due to information degeneracy. We present in the following experiments, intuitive approaches as baselines and our solution approach to mitigate this problem.

To systematically monitor the performance of different models, we introduce a metric that quantifies the total beta ratio errors relative to the correct parameters. This metric is defined as:
\begin{equation}\label{eq:error_beta}
E_{\beta}(\bm{\beta}) = \sum_{j\in\mathcal{J}}\left|\left|\frac{\beta_{MNL}^j}{\beta_{MNL}^{ref}} - \frac{\beta^j}{\beta^{ref}}\right|\right|  .
\end{equation}
Here, $\beta^{ref}$ serves as a reference variable for establishing the ratios. In our study, the parameter associated with the variable ‘Girl' is chosen as this reference.

\begin{table}[h!]
    \centering
    \begin{tabular}{rcc|ccc}\toprule
      Variable Ratios   & MNL($\mathcal{X}$) & L-MNL($\mathcal{X}$, $\mathcal{I}$) & MNL($\mathcal{X}'$) & L-MNL($\mathcal{X}',\mathcal{I}$) & L-MNL($\mathcal{X}',\mathcal{I}_p$) \\ \midrule
Intervention	&	0.26	&	--	&	0.33	&	--	&	--	\\
\rowcolor{gray!10}
CrossingSignal	&	0.48	&	0.53	&	0.63	&	0.80	&	0.48	\\
Man	&	0.61	&	0.53	&	--	&	--	&	--	\\
\rowcolor{gray!10}
Woman	&	0.72	&	0.55	&	0.52	&	0.46	&	0.71	\\
Pregnant	&	0.76	&	0.86	&	1.10	&	0.75	&	0.77	\\
\rowcolor{gray!10}
Stroller	&	0.79	&	0.94	&	1.15	&	0.89	&	0.80	\\
OldMan	&	0.31	&	0.24	&	$\textit{-0.05}$	&	$\textit{0.06}$	&	0.29	\\
\rowcolor{gray!10}
OldWoman	&	0.39	&	0.21	&	0.11	&	0.16	&	0.38	\\
Boy	&	0.89	&	0.94	&	0.86	&	0.83	&	0.88	\\
\rowcolor{gray!10}
Girl	&	1.00	&	1.00	&	1.00	&	1.00	&	1.00	\\
Homeless	&	0.41	&	0.33	&	0.33	&	0.26	&	0.40	\\
\rowcolor{gray!10}
LargeWoman	&	0.57	&	0.38	&	0.40	&	0.35	&	0.56	\\
LargeMan	&	0.43	&	0.30	&	0.18	&	0.16	&	0.41	\\
\rowcolor{gray!10}
Criminal	&	0.15	&	0.46	&	0.28	&	-0.21	&	0.14	\\
MaleExecutive	&	0.60	&	0.55	&	0.87	&	0.49	&	0.61	\\
\rowcolor{gray!10}
FemaleExecutive	&	0.71	&	0.55	&	0.92	&	0.58	&	0.71	\\
FemaleAthlete	&	0.78	&	0.53	&	0.63	&	0.56	&	0.77	\\
\rowcolor{gray!10}
MaleAthlete	&	0.66	&	0.48	&	0.40	&	0.41	&	0.64	\\
FemaleDoctor	&	0.73	&	0.69	&	1.00	&	0.69	&	0.74	\\
\rowcolor{gray!10}
MaleDoctor	&	0.65	&	0.70	&	1.00	&	0.60	&	0.65	\\
Dog	&	0.23	&	0.37	&	$\textit{0.07}$	&	0.23	&	0.21	\\
\rowcolor{gray!10}
Cat	&	0.17	&	0.27	&	$\textit{0.00}$	&	0.15	&	0.15	\\

         \midrule 
         Accuracy & $78.1\%$ & $79.4\%$ & 77.0$\%$ & $\bm{79.4}\%$ & 78.3$\%$ \\ \midrule
      $E_\beta$  &  0 & 2.62 & 4.25 & 2.87 & $\bm{0.18}$  \\   \bottomrule
    \end{tabular}
    \caption{Beta ratios standardized with $\beta_{ref}=$‘Girl'. On the left, there is the theroetical model where parameter weights are correct together with the biasing effect of an Image model with isomorphic data in $\mathcal{I}$. On the right are results when using incomplete specification $\mathcal{X'}$ and how complementary data from the images interact with the parameters. Numbers in italics do not pass the t-test for parameter significance.}
    \label{tab:1-1}
\end{table}

\subsection{Baselines Experiments}
The most intuitive approach to solving the issue of the NN model interfering with the correct statistical parameter weights in the case of data congruence is to control information flow via architecture and regularizers. While many methods have been tested, we present the most rational ones: the latent model to try to extract complementary information of the NN via a residual, or a $\beta$-VAE \citep{higgins2016beta} inspired model, to slice complimentary information out from the latent space. 
\label{sec:baselines}


\subsubsection{Latent Model}\label{sec:latent_model}
In the experiments detailed in Section \ref{sec:initial}, it was observed that the representation term derived from the image component of our L-MNL model interferes with the parameters of its interpretable counterpart. To try to regulate the flow of information and optimization of this representation term, and extract only the complementary information discernible from the image, we introduce three new elements to the base model.

Firstly, we incorporate a detection head into our latent space, which is trained to identify the values of $\mathcal{X}'_n$ using a Mean Square Error (MSE) loss, denoted as $L_D$. This head also generates an additional term, which acts as our residual term. Combined with the detection, this produces latent estimations $l_{xn}$, analogous to utility variables. Subsequently, we introduce a linear layer, $C(\cdot)$, that maps these detected values (along with the added residual term) to a secondary loss $L_{C2}$ for choice prediction. The parameter values within this layer, corresponding to their detected tabular inputs, can be interpreted as $\beta_{I}$ parameters for the detected image components.

To further refine our model, we introduce a regularization loss between $\beta_{I}$ and $\beta_{LMNL}$ to ensure their similarity. This step prevents $\beta_{I}$ from approaching zero which would allow the residual term to become the primary component for choice prediction. Ultimately, the residual term is the sole component channeled to the original utility function's loss, aiming to segregate the expression of correlated information present in both $\mathcal{X}_n$ and $\mathcal{I}_n$.

The introduced losses and the regularizer can be mathematically represented as:
\begin{align}
    \label{eq:detection_loss}
    \mathcal{L}_D(\mathcal{X}_n', \mathcal{I}_n) &= ||X_n' - D(B(\mathcal{I}_n))||^2 , \\
    \mathcal{L}_{C2}(y_i, \mathcal{I}_n) &=- \sum_{i\in\mathcal{C}} y_i\log(C(l_{xn})_i) , \\
     \mathcal{L}_\beta(\beta_{LMNL},\beta_I ) &= ||\beta_{LMNL} -\beta_I ||.
\end{align}
A visual representation of the model can be found in \ref{sec:models_appendix}. Controlling multiple losses requires tuning $\lambda$ hyper-parameters. In our case $\lambda_{C2}=\lambda_{\beta}=1$ and $\lambda_D=10$. \\

\subsubsection{$\beta$-VAE and Slicing}

The second baseline introduced in this paper draws inspiration from Variational Autoencoders (VAEs). VAEs are designed to reconstruct the original input after encoding it into a latent space. A particular variant, the $\beta$-VAE \citep{higgins2016beta,burgess2018understanding}, introduces a regularizing loss, \(L_{VAE}\), to ensure that the latent space is independent and identically distributed (i.i.d). The primary objective of this regularization is to compel each latent variable to capture as much of the reconstruction's information as possible, leading to distinct representations for each variable, also known as disentanglement.

In essence, if we can successfully regularize the latent space in this manner, it becomes feasible to ``slice" a portion of it to detect the original tabular inputs, $\mathcal{X}_n'$. The remaining portion should then encapsulate the complementary information. To achieve this, we define a bottleneck latent space, $l_{bn}$, derived from a linear layer, $B2(\cdot)$, such that $B2(B(\mathcal{I}_n)) = l_{bn}$. To apply the $\beta$-VAE regularization to this space, it is dimensioned as $L \times 2$, allowing us to estimate $L$ means, $\mu_{bn}$, and $L$ standard deviations, $\sigma_{bn}$.

Utilizing the reparametrization trick (as shown in equation \ref{eq:reparametrization}), we sample from this space and apply the $\beta$-VAE regularization (equation \ref{eq:beta_vae_regularization}). Subsequently, $d$ samples $z_{dn}$ are processed through a detection head, $D(\cdot)$, trained to detect $\mathcal{X}_n'$, while $r$ samples $z_{rn}$ pass through a linear head, $R(\cdot)$, to serve as a representation term added to the utility functions. It's crucial to note that $d + r = L$, and both $d$ and $r$ are assumed to be i.i.d representations.

The associated equations for this methodology are:
\begin{align}
    \label{eq:reparametrization}
    z_{bn} &= \mu_{bn} + \sigma_{bn} \odot \epsilon, \quad \text{where } \epsilon \sim \mathcal{N}(0,1), \\
    \label{eq:beta_vae_regularization}
    \mathcal{L}_{VAE} &= \frac{1}{2} \sum_{j=1}^{L} (1 + \log(\sigma_{bn}^2) - \mu_{bn}^2 - \sigma_{bn}^2), \\
    \label{eq:detection_loss_sample}
    \mathcal{L}_D &= ||\mathcal{X}_n' - D(z_{dn})||^2.
\end{align}
A figure of the model can be found in \ref{sec:models_appendix}. Optimal results are obtained for $\lambda_D=10$ and $\lambda_{VAE}=0.01$\\

\subsubsection{Baseline Results and Conclusions}

In our baseline experiments, the primary objective is to disentangle the complementary information present in $\mathcal{I}_n$ from that in $\mathcal{X}_n'$ by leveraging model architectures and various regularizers. However, as evidenced in Table \ref{tab:model_results}, the Neural Network component continues to introduce biases into the parameters of the DCM component. The underlying reason is that when the image component is tied to the prediction loss, it minimizes its loss by learning representations of significant variables from $\mathcal{X}_n'$. Given that it possesses internal representations of these variables, the interpretability constraint for L-MNL, as discussed in \citep{sifringer_enhancing_2020}, is inevitably compromised. Regularization and architectural designs do not suffice to mitigate these effects. Indeed, a key observation is that these baselines actually hold the highest accuracy and lowest loss compared to other models. This means that they make use of non-linear interactions of its variables that help with the choice prediction. To this extent, the values added to the utility function by the neural network still contain information related to the variables in the tabular data that we are trying to disentangle. 

This outcome might not be immediately intuitive. Therefore, we emphasize that when images are integrated with an interpretable DCM, any isomorphisms between $\mathcal{X}$ and $\mathcal{I}$ can influence the DCM parameters. Naturally, the frequency of such co-occurrences determines the extent of the bias introduced into the DCM component, as further explored in experiment \ref{sec:probabilistic}. Our proposed solution to this challenge is to eliminate these co-occurrences directly from the source image, a strategy we present in the subsequent section.

\begin{table}[h!]
    \centering
    \begin{tabular}{cccc}\toprule
        Model & Accuracy ($\%$) & $\mathcal{L}_{CCE}$ &  $E_{\beta}$ \\ \midrule
        MNL($\mathcal{X}'$) & 77.0 & 0.500 & 4.25 \\
        L-MNL($\mathcal{X}'$, $\mathcal{I}$) & $\bm{79.4}$ & $\bm{0.455}$ & 2.89\\
        L-MNL$_{latent}$($\mathcal{X}'$, $\mathcal{I}$) & 79.3 & 0.455 & 3.40 \\
        L-MNL$_{\beta VAE}$($\mathcal{X}'$, $\mathcal{I}$) & 79.2 & 0.458 & 2.58 \\
        L-MNL$_{s=100}$($\mathcal{X}'$, $\mathcal{I}_{b}$) & 78.2   & 0.474 & 0.41 \\
        L-MNL$_{s=100}$($\mathcal{X}'$, $\mathcal{I}_{p}$) & 78.3 & 0.486  & $\bm{0.181}$  \\ \midrule
        \textit{MNL($\mathcal{X}$)} & \textit{78.1} & \textit{0.486} & \textit{0} \\ \bottomrule
    \end{tabular}
    \caption{Model results for ratio error with respect to perfect model MNL($\mathcal{X}$).}
    \label{tab:model_results}
\end{table}

\subsection{Source Information Masking and Inpainting} \label{sec:exp_masking}

We now present our proposed approach, as outlined in section \ref{sec:methodology_mitigate}. While there are many ways to implement masking, the qualitative results depend on the end components and methods chosen for detection, saliency and finally masking. Given the synthetic nature of our study, we can actually rigorously examine the impact of any masking quality and thus the method's effectiveness for all settings.

\begin{figure}[ht]
    \centering
    \begin{minipage}[b]{0.49\linewidth}
        \centering
        \includegraphics[width=\linewidth]{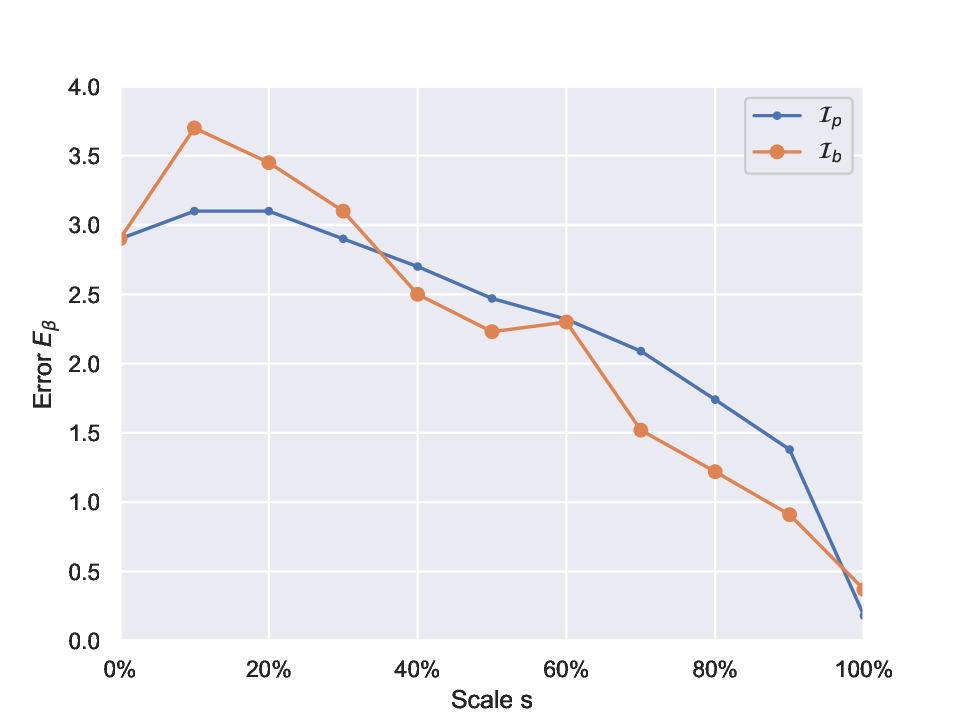}
        \caption{Masking $\mathcal{I}_b$ and inpainting $\mathcal{I}_p$ for different scales of s. The error $E_B$ on parameter estimates is greatly reduced as the masking potency increases.}
        \label{fig:masking_scale}
    \end{minipage}
    \hfill 
    \begin{minipage}[b]{0.49\linewidth}
        \centering
        \includegraphics[width=\linewidth]{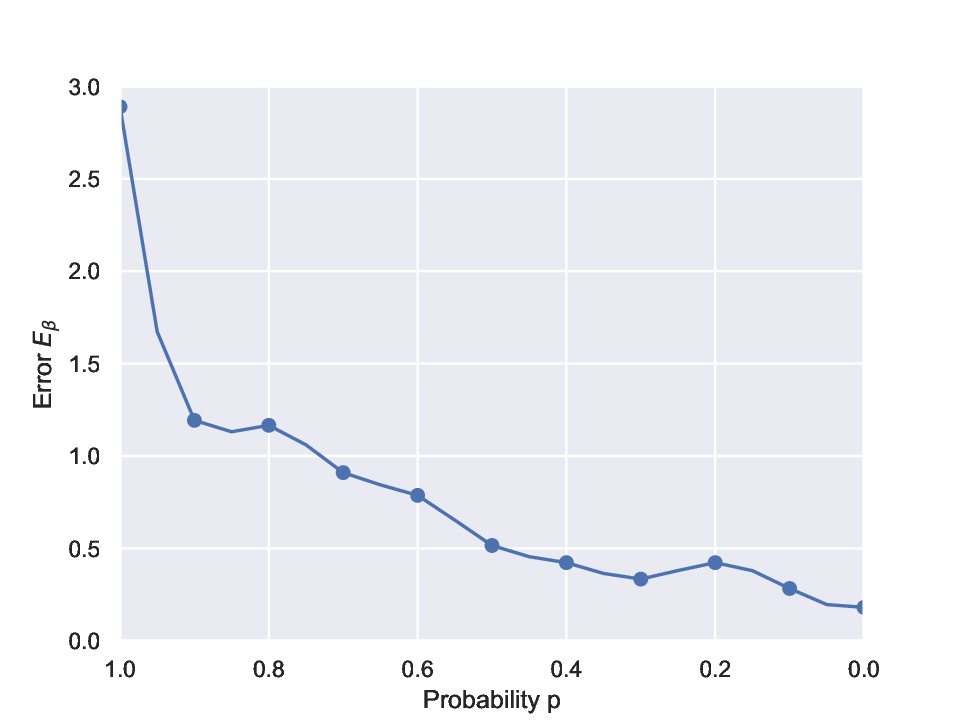}
        \caption{The summed error of parameter ratios given different probabilistic appearances of isomorphic information between both tabular and image datasets.}
        \label{fig:probabilistic}
    \end{minipage}
\end{figure}

A crucial aspect to consider is that the efficacy of this approach is closely tied to the precision of the masking step. If remnants of the representation elements from $\mathcal{X}_n'$ remain in the masked image $\mathcal{I}_{nb}$, the L-MNL model might still discern them during training time\footnote{Note that papers such as \cite{li2018tell} make use of this effect to algorithmically learn a more precise and fuller heatmap and masking step}. To thoroughly investigate these consequences, we simulate different qualities of masking by employing random black box masking on the character tiles. Similarly to the method shown in \cite{zhong2020random}, we scale the black box masking based on the $\%$ surface of the tile's total area. In this sense, a scale of $s=10\%$ would represent a poor quality of masking, while a scale of $s=100\%$ would represent perfect masking. The outcomes of L-MNL($\mathcal{X}_n'$, $\mathcal{I}_{nb}$) across various scales \(s\) are depicted in Figure (\ref{fig:masking_scale}).

It's worth noting that our experimental setup involves tiles of specific shapes. To emulate a more realistic scenario—where the model can't deduce the absent object based solely on its size—we introduce resizing augmentations. Additionally, to address the challenges posed by the multi-object detection nature of our experiment, we randomly add between 0 to 3 black boxes per choice. This ensures the model doesn't simply learn to count the concealed characters. The most favorable outcomes are achieved with a perfect mask, or \(s=100\%\), and this solution is reported in Table \ref{tab:model_results}.

As we may observe, the $\beta$ parameters are greatly fixed in comparison to the other methods. Ideally, an image that encapsulates only the complementary information should guide the model to exclusively learn this representation, thereby rectifying the parameters of the DCM component. However, the results suggest that the mere act of black boxing introduces enough perturbations to the input, preventing the model from achieving optimal performance.

To address this, we introduce the concept of inpainting. This ensures that the image, denoted as $\mathcal{I}_{np}$, remains devoid of informational gaps, presenting a consistent background throughout. In our experiments, we adopt a similar approach to random black box masking. However, instead of setting areas of character tiles to zero, we render them invisible. The outcomes of L-MNL($\mathcal{X}_{n}'$,$\mathcal{I}_{np}$) across various scales \(s\) can be viewed in Figure (\ref{fig:masking_scale}).

Notably, images at a scale of \(s=100\%\) exclusively showcase the missing complementary information from $\mathcal{X}_n'$. This allows the neural network component to learn its high-dimensional representation and incorporate it into the utility function without disrupting other parameters. As a result, the original statistical model's parameters are fully recovered, and the bias is fixed. This value is thus reported in Table (\ref{tab:model_results}).\\

\subsection{Probabilistic Congruence}\label{sec:probabilistic}

In our previous experiments, we operated under the assumption of a one-to-one isomorphism between each pair of tabular and image data entries. However, real-world datasets may not always exhibit such a direct correspondence. To address this, we extend our investigation to scenarios where isomorphic congruence between tabular and image data is probabilistic, rather than guaranteed.

We define a probability $p$ representing the likelihood of isomorphism within our dataset. A scenario where $p=0\%$ would be similar to the setting of L-MNL\(_{s=100}\)($\mathcal{X}_n'$, $\mathcal{I}_{np}$), indicating no isomorphism between the data types. Conversely, a scenario with $p=100\%$ aligns with L-MNL($\mathcal{X}_n'$, $\mathcal{I}_{n}$), where isomorphism is present in every data entry. The impact of varying probabilities of isomorphism on model performance is illustrated in Figure (\ref{fig:probabilistic}).

This probabilistic approach to congruence allows us to simulate a spectrum of real-world conditions, ranging from complete independence to full isomorphism between tabular and image data. The findings from this study provide critical insights into the robustness of parameter interpretability under varying degrees of data congruence. Indeed, when overlap in variables between both modalities appear at a low rate, interpretability of the statistical parameters are only slightly effected as there may be only very low biases.

\subsection{Real-World Applications and Limitations}\label{sec:real_world}

Our findings, while based on synthetic image manipulation, provide a valuable upper-bound benchmark for real-world applications involving 1-1 data modality isomorphism. In practical settings, where prediction accuracy is inherently lower, the interaction between the image model and the DCM component would likely be reduced, similar to what we observed in our masking and probabilistic congruence experiments.

Moreover, our study shows there are no clear and simple approaches in architecture modeling to avoid interference with the DCM interpretable parameters. Our research advocates for a proactive approach to information mitigation, which, while drastic, is necessary to avoid parameter bias. This methodology, as evidenced by our experiments with black box masking and inpainting, can be adapted to suit various real-world datasets.

Indeed, we can imagine, for instance, having access to a vacation house choice dataset, containing all valuable information in tabular form, and some pictures of the housing constituting the image data.  The images would contain valuable information such as the aesthetic appeal of the location and could be captured into a utility function, similar to \citep{van2020blending,van2023computer}. However, if the image model is able to detect the number of bathrooms throughout the images, this variable's parameter, if belonging to the utility function, would be compromised. Mitigating this issue would require detecting, and then masking or inpainting the elements that give away this information. The approach for such a dataset could range from masking detected bathroom pictures entirely, i.e., removing them from the data, or sampling image parts that render the detection of tabular information infeasible, yet the neural network may still learn about room quality and aesthetics.\\

\section{Conclusions and Future Research}\label{sec:conclusions}

In this study, we utilized the MIT moral machine dataset, a real-world choice experiment, alongside synthetic image reconstruction to benchmark a suite of experiments that delve into the intersection of tabular and image data. Our findings reveal that the inherent complexity of image data does not prevent a neural network from learning and reproducing redundant information when integrated with a Discrete Choice Modeling (DCM) framework and subjected to joint optimization. We have demonstrated that a neural network's capacity to encapsulate latent representations of isomorphic data undermines the interpretability constraints previously established in the literature \citep{sifringer_enhancing_2020}.\\

To address this, we have investigated two primary strategies: controlling the information flow within neural networks and masking source information. The former approach, despite its sophistication, proved insufficient in preventing the neural network's predictive capabilities from distorting the DCM's interpretable parameters. Consequently, we have established that mitigation of congruent information at the source is the most effective strategy. Techniques such as black masking and inpainting, which can be accessible via neural network saliency, have been shown to be effective in preserving the integrity of interpretable parameters with minimal neural network interference.

Moreover, we have extended our analysis to datasets where tabular and image data congruence is probabilistic, offering a nuanced understanding of how varying degrees of data overlap impact model performance.\\

Future work implies applying these methodologies to real-world datasets, which are slowly becoming available for Discrete Choice Modeling. As the integration of image data into DCM is still nascent, the availability of suitable datasets remains limited. Advancing this field will require not only innovative approaches to data integration but also the conception of these datasets that blend tabular and visual information in the domain of choice prediction.

\clearpage
\appendix

\section{Sanity Check}\label{sec:sanity}
We compare the predictive performance of both MNL and a simple 2-layer dense NN with those found in literature denoted $NN_w$ in Table (\ref{tab:sanity}) \citep{wiedeman2020modeling}. This is done on the same number of samples described in section (\ref{sec:data}), but without filtering out the different scenarios. We name this set $\mathcal{X}_{full}$.\\
As we may observe, we obtain similar accuracy for the standard statistical model as those seen in the literature. Moreover, the NN only performs slightly better than MNL, implying there are very few non-linear dependencies to be found given the tabular data. This works to our advantage, as it supposes that the original MNL model is not highly underfitting for the given data, and its interpretable parameters can therefore be trusted. \\
Now, we train the backbone Res18 B($\mathcal{X}_{full}$,$\mathcal{I}$) on the images in the task of both choice and tile detection thanks to linear heads $P(\cdot)$ and $D(\cdot)$. Results of both can be found in Table (\ref{tab:sanity}). We see that similar accuracy is obtained for choice by using only the image, and we reach practically $100\%$ accuracy in object detection. This is to our advantage for our correlation study, as this model with this synthetic data can act as an upper bound for any conclusions that arise concerning the ill effects of image and tabular data isomorphism. 

\begin{table}[h]
    \centering
    \begin{tabular}{ccc}\toprule
        Model & Accuracy ($\%$) & Detection ($\%$) \\ \midrule
        MNL($\mathcal{X}_{full}$) & 74 & -\\
        NN($\mathcal{X}_{full}$) & 75 & -\\
        $NN_w$($\mathcal{X}_{full}$) & 75 & -\\
        L-MNL($\mathcal{X}_{full}$, $\mathcal{I}$) & 75 & 100 \\ \bottomrule
    \end{tabular}
    \caption{Performance sanity check with literature, NN and $NN_w$ obtain similar results. Models with a detection head in this paper are able to reach 100$\%$ detection accuracy.}
    \label{tab:sanity}
\end{table}

\section{Model representations} \label{sec:models_appendix}

\begin{figure}[H]
    \centering
    \includegraphics[width=0.85\textwidth, trim={0 1.5cm 0 0},clip]{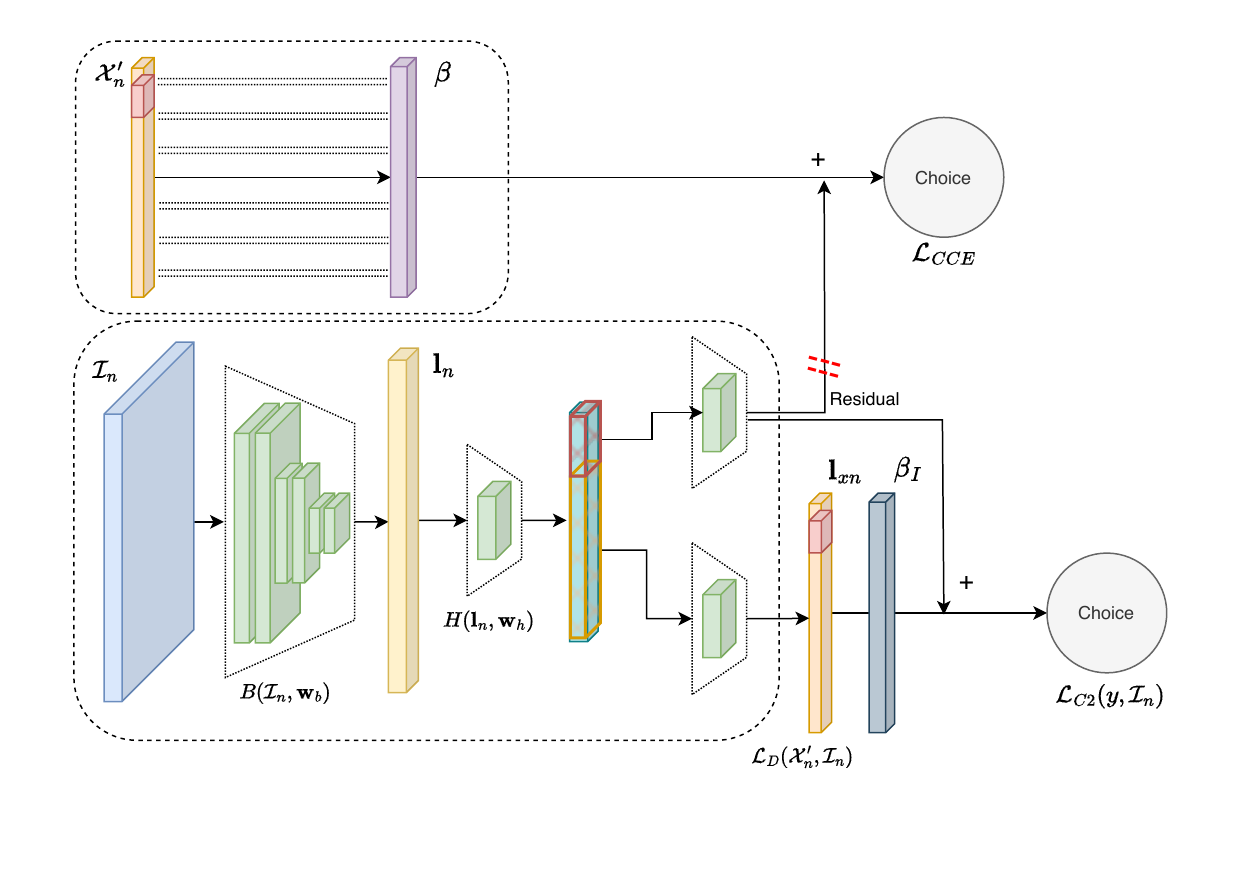}
    \caption{Latent Model representation. Two red lines indicate the gradient is not backpropagated. }
    \label{fig:latent_model}
\end{figure}

\begin{figure}[H]
    \centering
    \includegraphics[width=0.85\textwidth, trim={0 1.5cm 0 0},clip]{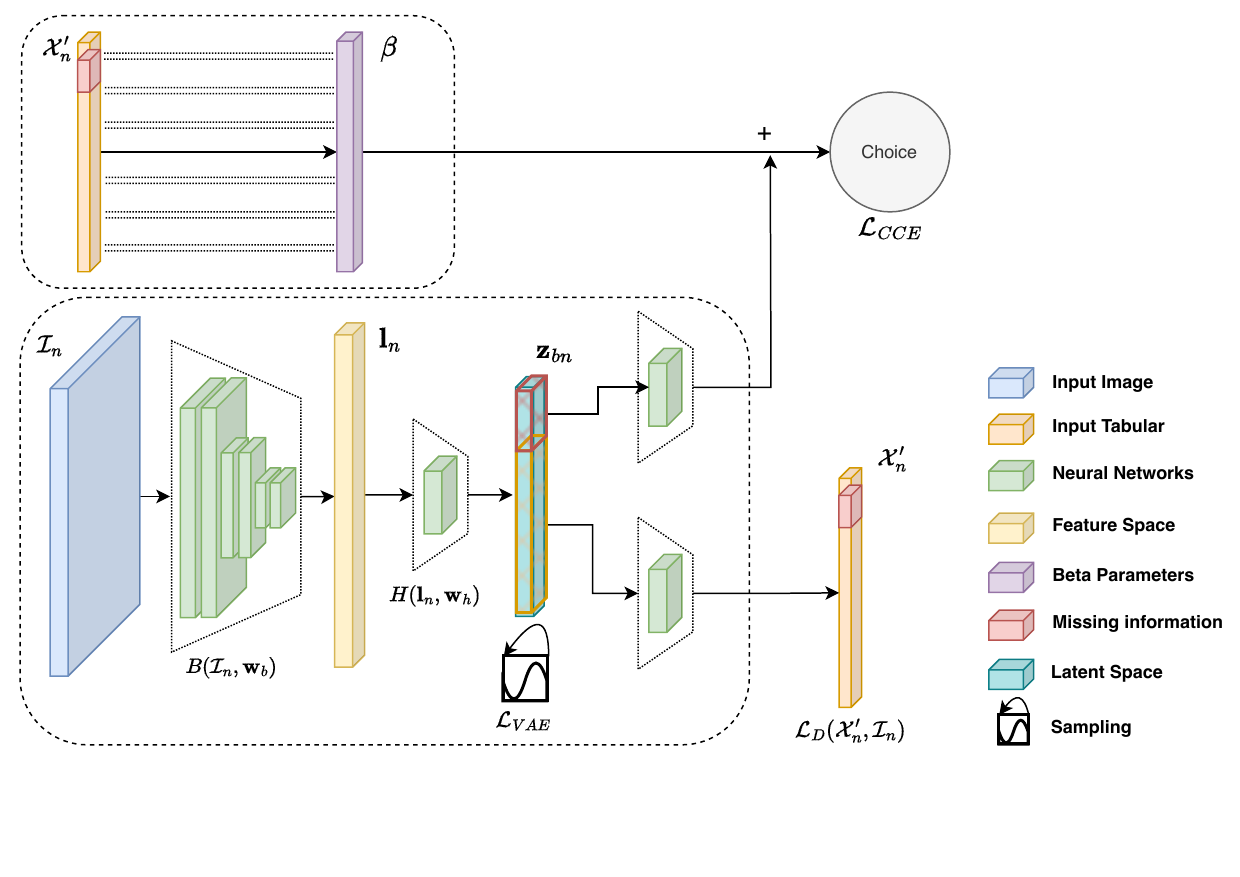}
    \caption{B-VAE inspired model representation. The samples are split for either detection or utility. }
    \label{fig:b-vae_model}
\end{figure}


\section*{Acknowledgments}
This work is supported by the Swiss
National Science Foundation under the grant $200021-
192326$. We also thank Edmond Awad and their team for sharing the additional resources from which we were able to create synthetic images.

{\small
\bibliography{main}
}

\end{document}